\documentclass[letterpaper]{article} % DO NOT CHANGE THIS
\usepackage{aaai2026}  % DO NOT CHANGE THIS
\usepackage{times}  % DO NOT CHANGE THIS
\usepackage{helvet}  % DO NOT CHANGE THIS
\usepackage{courier}  % DO NOT CHANGE THIS
\usepackage[hyphens]{url}  % DO NOT CHANGE THIS
\usepackage{graphicx} % DO NOT CHANGE THIS
\usepackage{natbib}  % DO NOT CHANGE THIS AND DO NOT ADD ANY OPTIONS TO IT
\usepackage{caption} % DO NOT CHANGE THIS AND DO NOT ADD ANY OPTIONS TO IT
\usepackage{algorithm}
\usepackage{algorithmic}

\usepackage{amssymb}
\usepackage{amsmath}
\usepackage{multirow}
\usepackage{booktabs}

\usepackage{times}
\usepackage{helvet}
\usepackage{courier}
\usepackage{xcolor}

\usepackage{newfloat}
\usepackage{listings}
\DeclareCaptionStyle{ruled}{labelfont=normalfont,labelsep=colon,strut=off} % DO NOT CHANGE THIS
\floatstyle{ruled}
\newfloat{listing}{tb}{lst}{}
\floatname{listing}{Listing}
\title{Training-Free Spatio-temporal Decoupled Reasoning Video Segmentation with Adaptive Object Memory}

\author{
    Zhengtong Zhu,
    Jiaqing Fan\thanks{Corresponding author.}, 
    Zhixuan Liu, 
    Fanzhang Li
}
\affiliations{
    School of Computer Science \& Technology, Soochow University, Suzhou, China\\
    20245227002@stu.suda.edu.cn, jqfan@suda.edu.cn, 2427405039@stu.suda.edu.cn, lfzh@suda.edu.cn
}

\usepackage{bibentry}
\begin{document}

\maketitle

\begin{abstract}
Reasoning Video Object Segmentation (ReasonVOS) is a challenging task that requires stable object segmentation across video sequences using implicit and complex textual inputs. Previous methods fine-tune Multimodal Large Language Models (MLLMs) to produce segmentation outputs, which demand substantial resources. Additionally, some existing methods are coupled in the processing of spatio-temporal information, which affects the temporal stability of the model to some extent. To address these issues, we propose Training-Free \textbf{S}patio-temporal \textbf{D}ecoupled Reasoning Video Segmentation with \textbf{A}daptive Object \textbf{M}emory (SDAM). We aim to design a training-free reasoning video segmentation framework that outperforms existing methods requiring fine-tuning, using only pre-trained models. Meanwhile, we propose an Adaptive Object Memory module that selects and memorizes key objects based on motion cues in different video sequences. Finally, we propose Spatio-temporal Decoupling for stable temporal propagation. In the spatial domain, we achieve precise localization and segmentation of target objects, while in the temporal domain, we leverage key object temporal information to drive stable cross-frame propagation. Our method achieves excellent results on five benchmark datasets, including Ref-YouTubeVOS, Ref-DAVIS17, MeViS, ReasonVOS, and ReVOS.
\end{abstract}

\begin{links}
    \link{Code}{https://github.com/machine928/SDAM}
\end{links}

 %================================Introduction================================%
\section{Introduction}
The ReasonVOS task was first proposed by VISA\cite{visa}, which differs from traditional Referring Video Object Segmentation (RefVOS) tasks \cite{mutr, dshmp, samwise, referdino}. It requires the stable segmentation of target objects in video sequences based on implicit and complex textual inputs, demanding advanced reasoning abilities. Due to the dynamic nature of video data, this task becomes even more challenging, where time-sensitive queries, occlusion, or rapid object movement may complicate the segmentation process.

% figure: introduction pipline
\begin{figure}[t]
\centering
\includegraphics[width=\columnwidth]{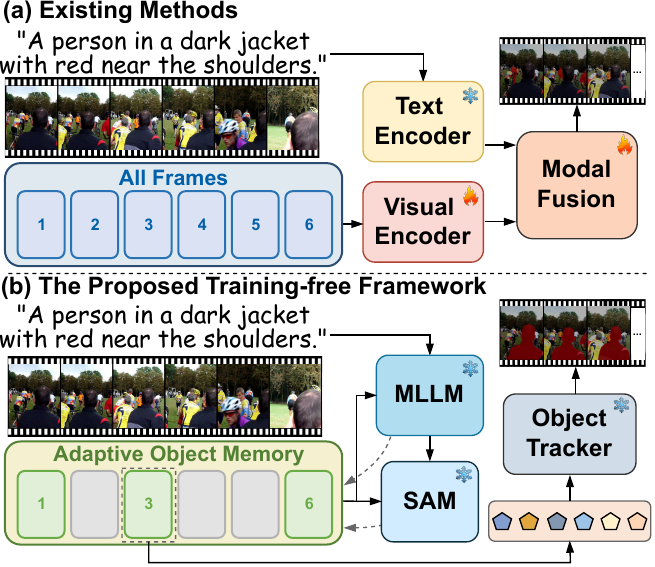}
\caption{Overview of SDAM. (a) Some existing methods rely on image-level video understanding, establishing correspondence between text descriptions and video frames, while neglecting the inherent temporal information in video tasks. (b) Our \textbf{training-free} framework adaptively memorizes key object information based on motion cues in the video and the frame-level confidence jointly obtained from MLLM and SAM. Additionally, we decouple the spatio-temporal information in the video to enhance the temporal stability of the architecture during the segmentation process.
}
\label{fig_introduction}
\end{figure}

The ReasonVOS task is an evolution of the RefVOS task, presenting a greater challenge. Previous methods based on MTTR \cite{mttr, referformer, vdit} primarily focus on image-level text-image correspondence, mainly segmenting objects in images by analyzing the given text descriptions. However, these methods suffer from coupling in the handling of spatio-temporal information and struggle to fully consider the temporal dependencies between frames, leading to difficulties in consistently producing accurate segmentation results in dynamic and complex scenes. For instance, we observe that in some videos on the Ref-YouTubeVOS dataset, the \(\mathcal{J\&F}\) metric experiences sharp declines in certain rapidly changing frames, which is caused by segmentation ambiguities arising from scene changes. Additionally, some methods attempt to propagate keyframe masks to achieve stable outputs across the entire video sequence \cite{find_track}. However, these methods rely on a fixed global sampling strategy to obtain keyframe candidates, which is challenging to account for the motion cues present in different video sequences. As a result, the spatio-temporal information contained in the candidate frame sequence is limited, making it difficult to obtain accurate keyframe. Moreover, to ensure good text-image correspondence, existing methods typically require significant time and computational resources for model training, this undoubtedly limits the development of the task.

To address these challenges, we propose SDAM, a novel framework based on MLLM with training-free characteristics (Figure \ref{fig_introduction}). Unlike previous methods, SDAM does not solely rely on traditional frame-level image understanding for segmentation. Instead, it combines motion cues and frame-level confidence in a spatio-temporal decoupling approach to consistently output masks. Specifically, our framework integrates existing pre-trained MLLMs \cite{qwen2.5}, image segmentation models \cite{sam}, and efficient object trackers \cite{cutie} to create a training-free reasoning video segmentation architecture. First, we analyze the motion information in the video sequence and adaptively select keyframe candidates, thereby obtaining a keyframe candidate set with richer spatio-temporal information (as shown in the Figure \ref{fig_introduction}). Then, using the outputs of MLLM and SAM, we compute confidence values for these candidate frames. The object with the highest confidence will be stored in the object memory bank for later use. Finally, with the help of the object tracker, we effectively track key objects throughout the video sequence, ensuring that the segmentation process remains stable and continuous across the entire video. 
% This training-free framework not only overcomes the dependency of traditional methods on large-scale training data and computational resources but also enhances the accuracy of inference video segmentation through Adaptive Object Memory and Spatio-temporal Decoupling mechanism.

To evaluate the effectiveness of our method, we conducted experiments on three RefVOS benchmark datasets (Ref-DAVIS 2017 \cite{ref_davis}, Ref-YouTube-VOS\cite{ref_ytb} and MeViS \cite{mevis}) and two ReasonVOS benchmark datasets (ReasonVOS \cite{videolisa} and ReVOS \cite{visa}). The quantitative and qualitative experimental results demonstrate the superiority of our approach. In summary, we highlight the following main contributions:
\begin{itemize}
    \item We propose a novel training-free reasoning video segmentation framework that surpasses existing fine-tuned methods by leveraging only pre-trained models.
    \item We adaptively select and memorize key objects by leveraging motion cues and frame-level confidence across the sequence, thereby capturing richer spatial information.
    \item We design this architecture from the perspective of spatio-temporal decoupling, first recognizing in the spatial domain and then propagating in the temporal domain, enhancing the temporal stability of the output.
    \item Our method achieves state-of-the-art results on five benchmark datasets. Specifically, we obtain \(\mathcal{J\&F}\) scores of \(\textbf{65.3}\%\), \(\textbf{76.0}\%\), \(\textbf{48.6}\%\), \(\textbf{55.1}\%\), and \(\textbf{58.0}\%\) on the Ref-YouTube-VOS, Ref-DAVIS17, MeViS, ReasonVOS, and ReVOS datasets, respectively.
\end{itemize}

%================================Related Work================================%
\section{Related Work}
\textbf{Referring Video Object Segmentation (RefVOS).} RefVOS aims to segment the target objects in a video based on a given explicit language expression. A recent benchmark, MeViS \cite{mevis} introduces complex multi-object scenes with extensive motion dynamics, further increasing the challenge of this task. Common approaches \cite{soc, tcdr, mutr, referdino, samwise} focus on attention mechanisms that use language queries to highlight the objects of interest. Additionally, some works \cite{mevis, dshmp, ssa, dmvs} have proposed motion aggregation techniques to capture motion information and methods like stable diffusion models to achieve modality fusion \cite{vdit}.

\subsubsection{Reasoning Segmentation.} Reasoning Segmentation \cite{lisa, glamm, gsva} has further advanced referring segmentation by generating masks from complex images and implicit text. LISA \cite{lisa} pioneers the field of inference segmentation. It introduces a new token to extend the vocabulary and proposes embedding as a masking paradigm that enhances segmentation to address scenarios requiring complex reasoning and world knowledge. PixelLM \cite{pixellm} integrates an innovative pixel decoder and a segmentation codebook with trainable tokens, enabling the efficient generation of high-quality masks without relying on external models. VISA \cite{visa} brings Reasoning Segmentation into the video domain. It fine-tunes pre-trained MLLM to select keyframes, segments them based on inference, and uses a pret-rained object tracker to propagate masks to other frames. However, it encounters challenges such as the limited representational capacity of a single specialized token and inaccurate keyframe selection, which hinder its segmentation and tracking performance. VRS-HQ \cite{vrshq} designs two special tokens and uses autoregressive learning of MLLM to effectively capture both local and global information.

%================================Method================================%
\section{Method}

\subsection{Overall Training-Free Pipeline}
\label{subsec:overall}
\begin{figure*}[t]
\centering
\includegraphics[width=\linewidth]{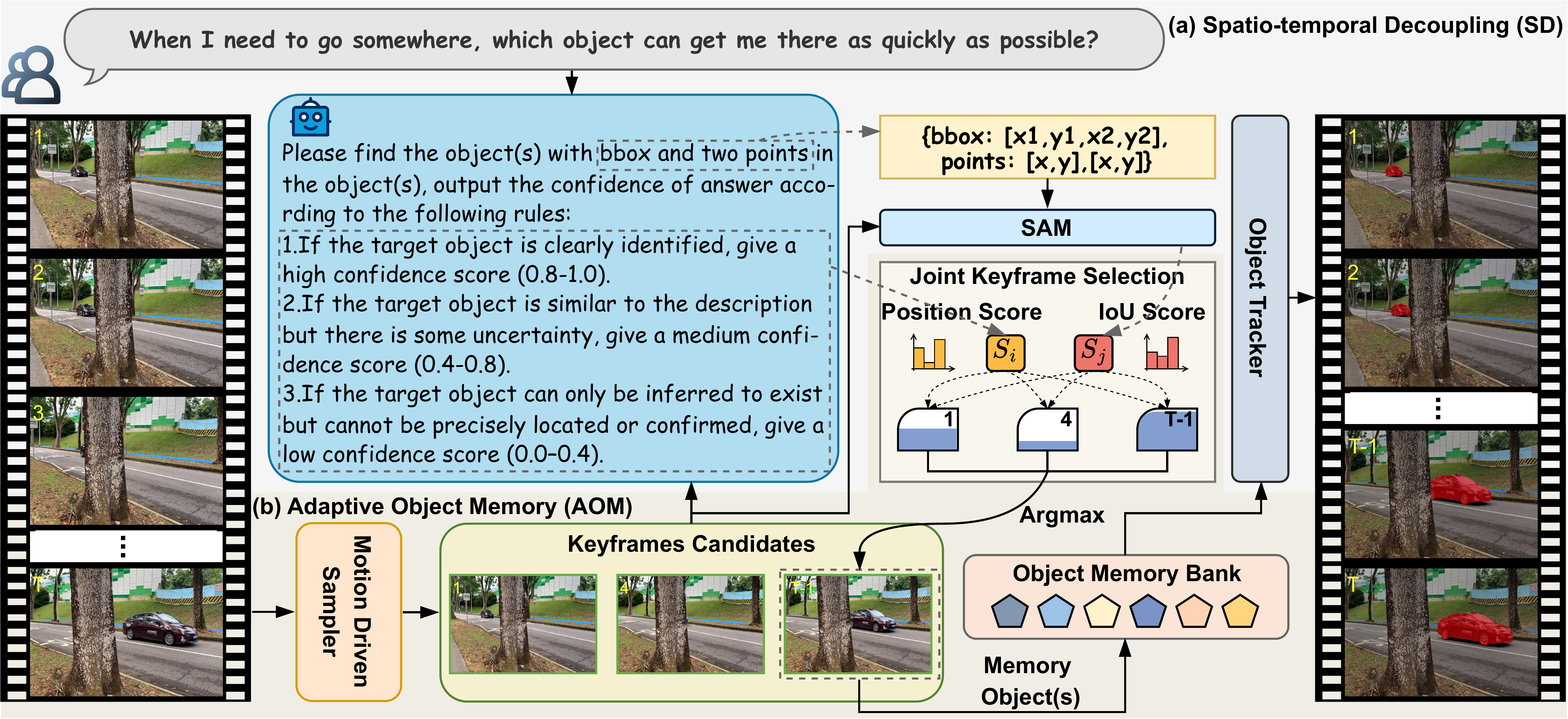}
\caption{The overall pipeline of SDAM. Our method consists of two parts: (a) Spatio-temporal Decoupling (SD). We pass the keyframe candidates into MLLM and SAM to obtain objects information in the spatial domain, and then use the Object Tracker to propagate the key object across the temporal domain. (b) Adaptive Object Memory (AOM). We first use Motion Driven Sampler to adaptively sample keyframe candidates based on motion cues, then use Joint Keyframe Selection to select the frame with the highest confidence as the keyframe, and store the key object memory in the Object Memory Bank.}
\label{fig_pipeline}
\end{figure*}

The overall pipeline of SDAM is shown in Figure \ref{fig_pipeline}. We integrate MLLM \cite{qwen2.5}, SAM \cite{sam}, and Cutie \cite{cutie} to form this framework, which mainly consists of two modules: AOM and SD. The AOM module is composed of the Motion Driven Sampler (MDS), Joint Keyframe Selection (JKS), and the Object Memory Bank (OMB). This module adaptively selects and memorizes key objects throughout the video sequence based on motion cues in the video. The SD module integrates MLLM, SAM and Cutie. In the spatial domain, we first input keyframe candidates into both MLLM and SAM. MLLM determines the location of the candidate objects in each frame, and then this location information is passed to SAM to obtain the object masks. In the temporal domain, we feed the outputs of MLLM and SAM into the AOM module to obtain the key object memory, then utilize the object tracking and segmentation capabilities of Cutie to achieve object segmentation across the entire video.

% figure: main pipeline

\subsubsection{Input and Output Settings.} Given a video sequence \(I = \left\{I_i \in \mathbb{R}^{H \times W \times 3}\right\}_{i=1}^T\) and a text query \(q\), where \(T\) is the length of the original video sequence, we input \(I\) into MDS to obtain keyframe candidates \(I^{\prime} = \left\{I_i \in \mathbb{R}^{H \times W \times 3}\right\}_{i=1}^{T^{\prime}}\), with \(T^{\prime}\) representing the length of the keyframe candidate sequence. Then, we input \(I^{\prime}\) and \(q\) into MLLM \(\mathcal{F}_M(\cdot)\) and input \(I^{\prime}\) into SAM \(\mathcal{F}_S(\cdot)\). Subsequently, we obtain the position information of the candidate objects:
\begin{equation}
    P = \mathcal{F}_M(I^{\prime},q) = \left\{P_i \in \mathbb{R}^N\right\}_{i=1}^{T^{\prime}},
    \label{eq:mllm_pos}
\end{equation}
where \(N\) represents the number of candidate objects in each frame, and \(P_i\) is output in JSON format. At the same time, \(\mathcal{F}_M(\cdot)\) also outputs the confidence scores \(S_{mllm} = \left\{S_i \in \mathbb{R}^1\right\}_{i=1}^{T^{\prime}}\) for each frame. After processing \(P\), it is input into \(\mathcal{F}_S(\cdot)\) to obtain the predicted masks for the candidate frames:
\begin{equation}
    M^{\prime} = \mathcal{F}_S(I^{\prime},P) = \left\{M_i \in \mathbb{R}^{H \times W \times N}\right\}_{i=1}^{T^{\prime}}.
    \label{eq:sam_mask}
\end{equation}
Additionally, the IOU scores for each frame \(S_{sam} = \left\{S_j \in \mathbb{R}^1\right\}_{j=1}^{T^{\prime}}\) are output. We then input \(S_{mllm}\) and \(S_{sam}\) into JKS to determine the keyframe \(I_{key} \in \mathbb{R}^{H \times W \times 3}\) and its mask \(M_{key} \in \mathbb{R}^{H \times W \times N}\), the key object memory \(O_{key}\) is obtained as follows:
\begin{equation}
    O_{key} = Memory(I_{key},M_{key}),
    \label{eq:obj_memory}
\end{equation}
where \(O_{key} \in \mathbb{R}^{N \times C}\), with \(C\) being the number of channels for object memory. We input \(O_{key}\) into the object memory bank \(O = \left\{O_i \in \mathbb{R}^{N \times C}\right\}_{i=1}^T\), then use the Object Tracker \(\mathcal{F}_{Obj}(\cdot)\) to propagate \(O_{key}\) across the entire video sequence, resulting in the final masks output:
\begin{equation}
    M = \mathcal{F}_{Obj}(I, O) = \left\{M_i \in \mathbb{R}^{H \times W \times N}\right\}_{i=1}^T.
    \label{eq:final_masks}
\end{equation}

\subsection{Adaptive Object Memory}
\label{subsec:am}
In this section, we will provide a detailed description of the design and implementation of the AOM module. Existing sampling-based methods \cite{find_track, think_video} typically adopt a global sampling strategy with fixed intervals. While this approach is simple to implement, it has significant limitations, especially when handling complex text queries. Global sampling may miss critical nodes in the video, particularly when specific actions mentioned in the text description occur at particular moments. For example, given the text query "bird stand on hand, then fly away," the action of "fly away" occurs only in a few frames of the video. If global sampling with fixed intervals is used, it may fail to accurately capture these important time points.

% figure: MDS
\begin{figure}[t]
\centering
\includegraphics[width=0.95\columnwidth]{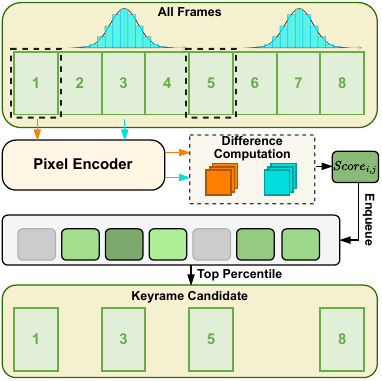}
\caption{Motion Driven Sampler. To obtain a keyframe candidate set with richer spatio-temporal information, we propose the MDS module, which can adaptively select frames with more significant scene changes as candidates based on the motion cues in the video sequence.}
\label{fig_mds}
\end{figure}

\subsubsection{Motion Driven Sampler.} To address the above issues, we propose an adaptive keyframe sampling strategy based on motion cues from the video sequence. By analyzing motion information and scene changes within the video, the MDS module can dynamically adjust the sampling strategy to select the frames that best reflect key actions. This adaptive sampling strategy not only improves the hit rate of keyframes but also enhances the spatio-temporal information of the sampling results, ensuring that the core events and actions described in the text are effectively captured.

The implementation details of MDS are shown in Figure \ref{fig_mds}. Given an existing video sequence \(I\), we first select \(m\) anchor points \(I^{a} = \left\{I_i^{a} \in \mathbb{R}^{H \times W \times 3}\right\}_{i=1}^m\) at fixed intervals within the video sequence. These anchor points divide the entire video sequence into \(m\) segments, each of length \(n\). Within these \(m\) segments, we need to calculate the degree of difference between the anchor frame and other frames in the sequence \(I^{o} = \left\{I_j^{o} \in \mathbb{R}^{H \times W \times 3}\right\}_{j=2}^n\), which serves as motion cues. The greater the difference, the more significant the motion change between \(I_i^{a}\) and \(I_j^{o}\). Specifically, we first input \(I_i^{a}\) and \(I_j^{o}\) into a pixel encoder \(\mathcal{F}_{pixelEnc}(\cdot)\) to obtain their respective feature maps \(F_i^{a} \in \mathbb{R}^{h \times w \times c}\) and \(F_j^{o} \in \mathbb{R}^{h \times w \times c}\). We then use \(F_i^{a}\) and \(F_j^{o}\) to calculate the difference metric \(D_{i,j}\) between \(I_i^{a}\) and \(I_j^{o}\), which effectively reflects the motion changes between the anchor frame and the sequence frame. This process is expressed as follows:
\begin{equation}
    F_i^{a},F_j^{o} = \mathcal{F}_{pixelEnc}(I_i^{a}),\mathcal{F}_{pixelEnc}(I_j^{o}),
    \label{eq:pixel_encode}
\end{equation}

\begin{equation}
    D_{i,j} = \mathcal{D}(F_i^{a},F_j^{o}), 
    \label{eq:feat_diff}
\end{equation}
where \(\mathcal{D}(\cdot)\) represents the difference metric function between feature maps. For the selection of candidate frames, we tend to choose frames with greater differences from the anchor frames, which means that our candidate sequence contains richer spatio-temporal information. However, due to the locality principle, the frames farther from the anchor point tend to have larger differences with the anchor, which inevitably leads to selecting frames that are farther away from the anchor point during sampling, i.e., frames closer to the next anchor point. This negatively impacts the diversity of the sampling. To address this issue, we use a normal distribution to balance this, and the score for each frame in the final subsequence is:
\begin{equation}
    Score_{i,j}=\exp \left(-\frac{\left(j-\frac{n}{2}\right)^2}{2 \sigma^2}\right) \cdot D_{i, j},
    \label{eq:candidate_score}
\end{equation}
where \(\sigma\) represents the standard deviation. We input the scores of all frames, except for the anchor frames, into a score queue and then select the frames corresponding to the top \(K\)-th percentile scores, along with the anchor frames, are output as the keyframe candidates \(I^{\prime}\).

\subsubsection{Joint Keyframe Selection.} To determine the frame in the keyframe candidates that best matches the text description, we combine the confidence outputs from MLLM and SAM as the basis for locating the keyframe. Suppose we have already obtained the confidence outputs \(S_{mllm}\) and \(S_{sam}\) from MLLM and SAM, respectively, using \(I^{\prime}\). We introduce a harmonizing parameter \(a\) to balance the weight of the two confidence values. The final score for the keyframe candidates \(S_{key} = \left\{S_k \in \mathbb{R}^1\right\}_{k=1}^{T^{\prime}}\) is calculated as follows:
\begin{equation}
    S_k = a \cdot S_{i} + (1 - a) \cdot S_{j},i=j=k,
    \label{eq:keyframe_score}
\end{equation}
where \(S_i \in S_{mllm}, S_j \in S_{sam}\), and \(a \in [0,1]\) is used to adjust the importance of \(S_i\) and \(S_j\). We select the frame corresponding to the highest score in \(S_{key}\) as the keyframe \(I_{key}\).

\subsubsection{Object Memory Bank.} We introduce OMB to store the high-dimensional semantic information of key objects for more precise object segmentation. Through this memory mechanism, we can continuously identify and track each object throughout the entire video sequence, thereby enhancing the stability of the segmentation output. Now that we have obtained the keyframe \(I_{key}\) and its mask \(M_{key}\), we utilize Cutie's \cite{cutie} object memory capability to implement memory storage for key objects (as in Equation \ref{eq:obj_memory}). It converts the keyframe and the corresponding segmentation mask into high-dimensional semantic features for storage, which are then used for the bidirectional propagation of object masks across the entire sequence, reducing segmentation errors caused by temporal variations.

% table: RefVOS comparison
\begin{table*}[t]
\centering
\begin{tabular}{@{}c|c|ccc|ccc|ccc@{}}
\toprule
\multirow{2}{*}{Method} & 
\multirow{2}{*}{Publication} & 
\multicolumn{3}{c|}{Ref-YouTube-VOS} & 
\multicolumn{3}{c|}{Ref-DAVIS17} & 
\multicolumn{3}{c}{MeViS} \\
 &  & \(\mathcal{J}\&\mathcal{F}\) & \(\mathcal{J}\) & \(\mathcal{F}\) & \(\mathcal{J}\&F\) & \(\mathcal{J}\) & \(\mathcal{F}\) & \(\mathcal{J}\&F\) & \(\mathcal{J}\) & \(\mathcal{F}\) \\ \midrule
LBDT \cite{lbdt} & CVPR 2022 & 49.4 & 48.2 & 50.6 & 54.5 & - & - & 29.3 & 27.8 & 30.8 \\
ReferFormer \cite{referformer} & CVPR 2022 & 62.9 & 61.3 & 64.6 & 61.1 & 58.1 & 64.1 & 31.0 & 29.8 & 32.2 \\
VLT+TC \cite{vlt} & TPAMI 2023 & 62.7 & - & - & 60.3 & - & - & 35.6 & 33.6 & 37.3 \\
HTML\cite{html} & ICCV 2023 & 63.4 & 61.5 & 65.2 & 62.1 & 59.2 & 65.1 & - & - & - \\
OnlineRefer \cite{online_refer} & ICCV 2023 & 63.5 & 61.6 & 65.5 & 64.8 & 61.6 & 67.7 & - & - & - \\
LISA \cite{lisa} & CVPR 2024 & 54.4 & 54.0 & 54.8 & 66.0 & 63.2 & 68.8 & 37.9 & 35.8 & 40.0 \\
VISA \cite{visa} & ECCV 2024 & 63.0 & 61.4 & 64.7 & \underline{70.4} & \underline{67.0} & \underline{73.8} & 44.5 & 41.8 & 47.1 \\
VideoLISA \cite{videolisa} & NeurIPS 2024 & 63.7 & 61.7 & 65.7 & 68.8 & 64.9 & 72.7 & 44.4 & 41.3 & 47.6 \\
DMVS \cite{dmvs} & CVPR 2025 & \underline{64.3} & \underline{62.4} & 66.2 & 65.2 & 62.2 & 68.2 & \textbf{48.6} & \underline{44.2} & \underline{52.9} \\
SSA \cite{ssa} & CVPR 2025 & \underline{64.3} & 62.2 & \underline{66.4} & 67.3 & 64.0 & 70.7 & \textbf{48.6} & 44.0 & \textbf{53.2} \\
\midrule
\textbf{SDAM} & Ours & \textbf{65.3} & \textbf{63.4} & \textbf{67.1} & \textbf{76.0} & \textbf{73.2} & \textbf{78.8} & \textbf{48.6} & \textbf{45.5} & 51.7 \\ 
\bottomrule
\end{tabular}
\caption{Performance comparison with previous methods on the validation sets of RefVOS datasets. The highest result in each column will be marked in \textbf{bold}, and the second-highest result will be \underline{underlined}.
}
\label{tab:ref_vos_comparison}
\end{table*}

\subsection{Spatio-temporal Decoupling}
\label{subsec:sd}
Previous methods based on MTTR \cite{referformer, mutr, vdit} have achieved good results, but these architectures are limited to frame-level image understanding, thus ignoring the inherent temporal information in video sequences. This can lead to ambiguity in segmentation when objects in the video sequence resemble the appearance or behavior of the referenced object, causing the model to struggle in consistently outputting the target object's mask. For example, given the same video, if the text queries \(q_1\): "a giraffe is walking on the grass field from the left" and \(q_2\): "a giraffe walking rightwards towards another on the right of the view" are provided, and only frame-level image understanding is used, when the two giraffes switch positions, the model may mistakenly output masks for mismatched instances based on the positional information in the text description. Therefore, we propose the SD mechanism, which uses MLLM \cite{qwen2.5} and SAM \cite{sam} to locate and segment the target objects, and then employs Cutie \cite{cutie} to propagate the target object masks across the temporal domain.

\subsubsection{Confidence-based Key Object Segmentation.} In the spatial domain, we first utilize the real-world reasoning ability of \(\mathcal{F}_M(\cdot)\) to output candidate key object location information in JSON format. This is then passed into \(\mathcal{F}_S(\cdot)\) to obtain the candidate key object mask \(M^{\prime}\) (as in Equations \ref{eq:mllm_pos} and \ref{eq:sam_mask}). The accuracy with which we can locate the object mask most closely matching the text description from the keyframe candidates directly determines the segmentation result of the entire video sequence, which requires the confidence output to be as precise as possible. We use the IOU score of the mask output by \(\mathcal{F}_S(\cdot)\) as the segmentation confidence \(S_{sam}\), and the output of \(\mathcal{F}_M(\cdot)\) as the object confidence \(S_{mllm}\).

During the experiments, we found that the accuracy of \(S_{mllm}\) has a significant impact on the final segmentation result. However, we observed that MLLM, under a broad prompt setting for "output the confidence of answer," has difficulty providing accurate scores. Therefore, we combined its reasoning capability to design a confidence-level output prompt (as shown in Figure \ref{fig_pipeline}). Additionally, we also require MLLM to output its reasoning process to guide the model in selecting the appropriate confidence intervals.

\subsubsection{Temporal Propagation.} Now that we have obtained the key object mask \(M_{key}\) in the spatial domain, our next goal is to stabilize the tracking and segmentation of the target object throughout the entire video sequence. We need to leverage the temporal coherence of the object and the visual context in adjacent frames to ensure accurate segmentation across the entire sequence. To achieve this, we introduced the efficient semi-supervised VOS method, Cutie, as our object tracker. We divide the entire video sequence into two parts based on the position of the keyframe: \(I^{fw} = \left\{I_{key}, I_{key-1}, \ldots, I_1\right\}\) and \(I^{bw} = \left\{I_{key}, I_{key+1}, \ldots, I_T\right\}\). Then, \(\mathcal{F}_{Obj}\) uses the dynamically maintained object memory bank \(O\) to bidirectionally propagate the key object mask, starting from the keyframe, and ultimately obtain the mask output for each frame:
\begin{equation}
    M_i= \begin{cases}F_{Obj}\left(I^{fw}, O\right), 1 \leqslant i < key \\ F_{Obj}\left(I^{bw}, O\right), \text{key} \leqslant i \leqslant T.\end{cases}
    \label{eq:bi_propagate}
\end{equation}
Since the presented solution is training-free, it does not rely on any loss function for model optimization.

%table: ReVOS comparison
\begin{table*}[t]
\centering
\begin{tabular}{ccccccccccc}
\toprule
\multicolumn{1}{c|}{\multirow{2}{*}{Method}} & \multicolumn{1}{c|}{\multirow{2}{*}{Publication}} & \multicolumn{3}{c|}{Referring} & \multicolumn{3}{c|}{Reasoning} & \multicolumn{3}{c}{Overall} \\
\multicolumn{1}{c|}{} & \multicolumn{1}{c|}{} & \(\mathcal{J}\&F\) & \(\mathcal{J}\) & \multicolumn{1}{c|}{\(\mathcal{F}\)} & \(\mathcal{J}\&F\) & \(\mathcal{J}\) & \multicolumn{1}{c|}{\(\mathcal{F}\)} & \(\mathcal{J}\&F\) & \(\mathcal{J}\) & \(\mathcal{F}\) \\ \midrule
\multicolumn{11}{c}{\textit{Traditional methods without reasoning ability}} \\ \midrule
\multicolumn{1}{c|}{ReferFormer \cite{referformer}} & \multicolumn{1}{c|}{CVPR 2022} & 32.7 & 31.2 & \multicolumn{1}{c|}{34.3} & 23.4 & 21.3 & \multicolumn{1}{c|}{25.6} & 28.1 & 26.2 & 29.9 \\
\multicolumn{1}{c|}{LMPM \cite{mevis}} & \multicolumn{1}{c|}{ICCV2023} & 34.1 & 29.0 & \multicolumn{1}{c|}{39.1} & 18.8 & 13.3 & \multicolumn{1}{c|}{24.3} & 26.4 & 21.2 & 31.7 \\ \midrule
\multicolumn{11}{c}{\textit{LLM-based methods with reasoning ability}} \\ \midrule
\multicolumn{1}{c|}{TrackGPT-7B \cite{track_gpt}} & \multicolumn{1}{c|}{arXiv 2023} & 48.2 & 46.7 & \multicolumn{1}{c|}{49.7} & 39.0 & 36.8 & \multicolumn{1}{c|}{41.2} & 43.6 & 41.8 & 45.5 \\
\multicolumn{1}{c|}{TrackGPT-13B \cite{track_gpt}} & \multicolumn{1}{c|}{arXiv 2023} & 49.5 & 48.3 & \multicolumn{1}{c|}{50.6} & 40.5 & 38.1 & \multicolumn{1}{c|}{42.9} & 45.0 & 43.2 & 46.8 \\
\multicolumn{1}{c|}{LISA-7B \cite{lisa}} & \multicolumn{1}{c|}{CVPR 2024} & 45.7 & 44.3 & \multicolumn{1}{c|}{47.1} & 36.1 & 33.8 & \multicolumn{1}{c|}{38.4} & 40.9 & 39.1 & 42.7 \\
\multicolumn{1}{c|}{LISA-13B \cite{lisa}} & \multicolumn{1}{c|}{CVPR 2024} & 46.6 & 45.2 & \multicolumn{1}{c|}{47.9} & 36.7 & 34.3 & \multicolumn{1}{c|}{39.1} & 41.6 & 39.8 & 43.5 \\
\multicolumn{1}{c|}{VISA-7B \cite{visa}} & \multicolumn{1}{c|}{ECCV 2024} & 50.9 & 49.2 & \multicolumn{1}{c|}{52.6} & 43.0 & 40.6 & \multicolumn{1}{c|}{45.4} & 46.9 & 44.9 & 49.0 \\
\multicolumn{1}{c|}{VISA-13B \cite{visa}} & \multicolumn{1}{c|}{ECCV 2024} & 57.4 & 55.6 & \multicolumn{1}{c|}{59.1} & 44.3 & 42.0 & \multicolumn{1}{c|}{46.7} & 50.9 & 48.8 & 52.9 \\
\multicolumn{1}{c|}{GLUS \cite{glus}} & \multicolumn{1}{c|}{CVPR 2025} & \underline{58.3} & \underline{56.0} & \multicolumn{1}{c|}{\underline{60.7}} & 51.4 & 48.8 &  \multicolumn{1}{c|}{53.9} & \underline{54.9} & \underline{52.4} & \underline{57.3} \\
\multicolumn{1}{c|}{InstructSeg \cite{instruct_seg}} & \multicolumn{1}{c|}{ICCV 2025} & 57.0 & 54.8 & \multicolumn{1}{c|}{59.2} & \underline{51.9} & \underline{49.2} &  \multicolumn{1}{c|}{\underline{54.7}} & 54.5 & 52.0 & 56.9 \\ \midrule
\multicolumn{1}{c|}{\textbf{SDAM}} & \multicolumn{1}{c|}{Ours} &  \textbf{61.5} & \textbf{59.0} & \multicolumn{1}{c|}{\textbf{63.9}} & \textbf{54.6} & \textbf{51.8} & \multicolumn{1}{c|}{\textbf{57.5}} & \textbf{58.0} & \textbf{55.4} & \textbf{60.7} \\ \bottomrule
\end{tabular}
\caption{Performance comparison with previous methods on the validation set of ReVOS dataset.}
\label{tab:revos_comparison}
\end{table*}

% table: reasonVOS
\begin{table}[t]
\centering
\normalsize
\begin{tabular}{@{}ccccc@{}}
\toprule
\multicolumn{1}{c|}{\multirow{2}{*}{Method}} & \multicolumn{1}{c|}{\multirow{2}{*}{Publication}} & \multicolumn{3}{c}{ReasonVOS} \\
\multicolumn{1}{c|}{} & \multicolumn{1}{c|}{} & \(\mathcal{J}\&F\) & \(\mathcal{J}\) & \(\mathcal{F}\) \\ \midrule
\multicolumn{5}{c}{\textit{Traditional methods without reasoning ability}} \\ \midrule
% \multicolumn{1}{c|}{
% \begin{tabular}[c]{@{}c@{}} ReferFormer \\ \cite{referformer} \end{tabular}} & \multicolumn{1}{c|}{CVPR'22} & 32.9 & 30.2 & 35.6 \\ \midrule
\multicolumn{1}{c|}{
\begin{tabular}[c]{@{}c@{}} SOC \\ \cite{soc} \end{tabular}} & \multicolumn{1}{c|}{NeurIPS'23} & 35.9 & 33.3 & 38.5 \\ \midrule
\multicolumn{1}{c|}{
\begin{tabular}[c]{@{}c@{}} SgMg \\ \cite{sgmg} \end{tabular}} & \multicolumn{1}{c|}{ICCV'23} & 36.2 & 33.7 & 38.7 \\ \midrule
\multicolumn{1}{c|}{
\begin{tabular}[c]{@{}c@{}} OnlineRefer \\ \cite{online_refer} \end{tabular}} & \multicolumn{1}{c|}{ICCV'23} & 38.7 & 34.6 & 42.9 \\ \midrule
\multicolumn{5}{c}{\textit{LLM-based methods with reasoning ability}} \\ \midrule

\multicolumn{1}{c|}{
\begin{tabular}[c]{@{}c@{}} LISA \\ \cite{lisa} \end{tabular}} & \multicolumn{1}{c|}{CVPR'24} & 31.1 & 29.1 & 33.1 \\ \midrule

\multicolumn{1}{c|}{
\begin{tabular}[c]{@{}c@{}} VideoLISA \\ \cite{videolisa} \end{tabular}} & \multicolumn{1}{c|}{NeurIPS'24} & 47.5 & 45.1 & 49.9 \\ \midrule

\multicolumn{1}{c|}{
\begin{tabular}[c]{@{}c@{}} GLUS \\ \cite{glus} \end{tabular}} & \multicolumn{1}{c|}{CVPR'25} & \underline{49.9} & \underline{47.5} & \underline{52.4} \\ \midrule

\multicolumn{1}{c|}{\textbf{SDAM}} & \multicolumn{1}{c|}{Ours} & \textbf{55.1} & \textbf{51.3} & \textbf{58.8} \\ \bottomrule
\end{tabular}
\caption{Performance comparison with previous methods on the ResonVOS dataset.}
\label{tab:reason_vos_comparison}
\end{table}

%table: sampling strategy
\begin{table}[t]
\centering
\normalsize
\begin{tabular}{c|ccc|c}
\toprule
\multirow{2}{*}{Sampling Strategy} & \multicolumn{3}{c|}{Ref-DAVIS17} & \(\text{Val}_u\) \\
 & \(\mathcal{J\&F}\) & \(\mathcal{J}\) & \(\mathcal{F}\) & \(\mathcal{J\&F}\) \\\midrule
First Frame & 69.2 & 66.3 & 72.1 & 50.6 \\
Global & 73.4 & 70.8 & 75.9 & 54.4 \\
\small Motion Driven Sampler & \textbf{76.0} & \textbf{73.2} & \textbf{78.8} & \textbf{55.8} \\ \bottomrule
\end{tabular}
\caption{Ablation study on sampling strategies.}
\label{tab:sampling}
\end{table}

%table: confidence_a
\begin{table}[t]
\centering
\begin{tabular}{cc|ccc|c}
\toprule
\multicolumn{2}{c|}{\begin{tabular}[c]{@{}c@{}}Confidence\\  Weight\end{tabular}} & \multicolumn{3}{c|}{Ref-DAVIS17} & ReasonVOS \\
\(a\) & \(1-a\) & \(\mathcal{J\&F}\) & \(\mathcal{J}\) & \(\mathcal{F}\) & \(\mathcal{J\&F}\) \\ \midrule
0.4 & 0.6 & 73.9 & 71.4 & 76.4 & 54.1 \\
0.5 & 0.5 & 74.7 & 72.3 & 77.1 & 54.8 \\
0.6 & 0.4 & 75.8 & 73.1 & 78.5 & \textbf{55.1} \\
0.75 & 0.25 & \textbf{76.0} & \textbf{73.2} & \textbf{78.8} & 54.7 \\ \bottomrule
\end{tabular}
\caption{Ablation study on confidence weight.}
\label{tab:confidence_a}
\end{table}

\section{Experiments}

% \subsection{Implementation Details}
% We use the fine-tuned Qwen2.5-VL-7B \cite{vision_reasoner} as the inference model in the framework, while using SAM \cite{sam} as the segmentation model and leveraging Cutie \cite{cutie} to propagate key objects across the video sequence. Our method does not require any additional training, which ensures that it does not consume too many resources. Nevertheless, due to the large memory requirements of MLLM, our experiments are conducted on a single A100 GPU with 40GB of memory.

\subsection{Quantitative Results}

\subsubsection{Ref-YouTube-VOS.} As shown in Table \ref{tab:ref_vos_comparison}, SDAM, as a training-free architecture, surpasses existing state-of-the-art methods that require training on the Ref-YouTube-VOS \cite{ref_ytb} dataset. Specifically, our method achieves a \(\mathcal{J\&F}\) score of \(65.3\%\) on Ref-YouTube-VOS, leading the previous state-of-the-art methods DMVS\cite{dmvs} and SSA\cite{ssa} by \(1\%\) on this metric. SDAM demonstrates a certain advantage on this dataset.

\subsubsection{Ref-DAVIS17.} SDAM significantly outperforms existing state-of-the-art methods on the Ref-DAVIS17 \cite{ref_davis} dataset, achieving a \(\mathcal{J\&F}\) score of \(76.0\%\), leading the second-ranked method VISA \cite{visa} by \(5.6\%\), and the third-ranked method VideoLISA \cite{videolisa} by \(7.2\%\). SDAM demonstrates exceptional competitiveness on the Ref-DAVIS17 dataset. Due to the presence of similar objects belonging to the same category in multiple videos of the Ref-DAVIS17 dataset, some previous methods struggle to produce stable results. However, SDAM cleverly utilizes the SD mechanism to enhance the temporal stability of the output, significantly improving the model's performance on this dataset.

\subsubsection{MeViS.} MeViS \cite{mevis}, as a recently emerging benchmark dataset in the RefVOS field, is highly challenging due to its introduction of complex multi-object scenes with extensive motion. Our SDAM achieves a score of \(48.6\%\) in \(\mathcal{J\&F}\) on this dataset, tying for first place with the existing state-of-the-art methods DMVS and SSA, while leading these methods by \(1.3\%\) and \(1.5\%\) in the \(\mathcal{J}\) metric, respectively. Notably, DMVS and SSA are models specifically designed for MeViS. This demonstrates the robustness of our method when facing complex dynamic scenes.

\subsubsection{ReVOS.} As shown in Table \ref{tab:revos_comparison}, we significantly outperform existing state-of-the-art methods on all metrics in the large-scale Reasoning-based ReVOS \cite{visa} dataset. Our method achieves \(\mathcal{J\&F}\) scores of \(61.5\%\), \(54.6\%\), and \(58.0\%\) in the "Referring", "Reasoning" and "Overall" categories, respectively, surpassing the second-place methods by \(3.2\%\), \(2.7\%\), and \(3.1\%\).The experimental result shows that our method demonstrates outstanding performance in implicit text and complex multi-object scenarios.

\subsubsection{ReasonVOS.} As shown in Table \ref{tab:reason_vos_comparison}, we evaluate our method on the ReasonVOS \cite{videolisa} dataset. SDAM achieves significant results on this challenging dataset, with a \(\mathcal{J\&F}\) score of \(55.1\%\), surpassing the existing state-of-the-art method GLUS \cite{glus} by \(5.2\%\).

\subsection{Ablation Studies}

\subsubsection{Sampling Strategy.} As shown in Table \ref{tab:sampling}, we conduct ablation experiments on three sampling strategies on the Ref-DAVIS17 and MeVis (\(\text{Val}_u\))dataset. The anchor frame settings in MDS are the same as the global sampling interval, both set to \(\left\lfloor\frac{T}{4}\right\rfloor\), and this setting is used for the anchor frames in subsequent experiments. Building upon global sampling, we use MDS to improve the \(\mathcal{J\&F}\) metric from \(73.4\%\) to \(76.0\%\) on Ref-DAVIS17, and from \(54.4\%\) to \(55.8\%\) on MeViS (\(\text{Val}_u\)). Experimental result shows that our MDS can obtain keyframe candidates with richer spatio-temporal information, thereby improving the keyframe hit rate.

\subsubsection{Confidence Weight.} As shown in Table \ref{tab:confidence_a}, we conduct experiments on the confidence weight \(a\) based on MDS. We find that when \(a\) is set to 0.75, the best performance of \(76.0\% \ \mathcal{J\&F}\) is achieved on the Ref-DAVIS17 dataset, and when \(a\) is set to 0.6, the best performance of \(55.1\%\mathcal{J\&F}\) is achieved on the ReasonVOS dataset. The experimental results suggest that the final performance of the model is positively correlated with the weight of \(S_{mllm}\).

\subsubsection{Number of Keyframes.} We hypothesize that selecting a single keyframe might lead to incorrect masks for the entire video sequence due to errors in keyframe selection. To enhance the robustness of the method, we try selecting the top \(n\) keyframes with the highest confidence and perform bidirectional propagation from multiple keyframe nodes, thus avoiding error propagation caused by mismatching a single keyframe. We conduct experiments on the Ref-DAVIS17 dataset with the global sampling mode, and the experimental results are shown in Table \ref{tab:keyframe}. When \(n=1\), we achieve the optimal result of \(73.4\%\mathcal{J\&F}\), which proves that a single keyframe remains the best choice.

% figure: temporal analysis
\begin{figure}[t]
\centering
\includegraphics[width=\linewidth]{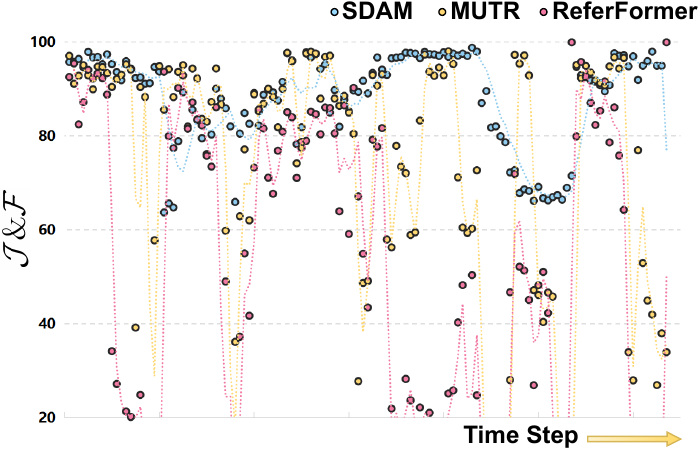}
\caption{Visualization of temporal stability analysis. We analyze the temporal stability of SDAM, MUTR, and ReferFormer on the Ref-YouTube-VOS dataset.}
\label{fig_temporal_ana}
\end{figure}

%table: keyframe
\begin{table}[t]
\centering
\begin{tabular}{c|ccc}
\toprule
\multirow{2}{*}{Number of Keyframes} & \multicolumn{3}{c}{Ref-DAVIS17} \\
 & \(\mathcal{J\&F}\) & \(\mathcal{J}\) & \(\mathcal{F}\) \\ \midrule
\(n=3\) & 68.6 & 66.3 & 70.9 \\
\(n=2\) & 70.7 & 69.4 & 72.0 \\
\(n=1\) & \textbf{73.4} & \textbf{70.8} & \textbf{75.9} \\ \midrule
\end{tabular}
\caption{Ablation study on the number of keyframes.}
\label{tab:keyframe}
\end{table}

\subsection{Temporal Stability Analysis}
As shown in Figure \ref{fig_temporal_ana}, we analyze the temporal stability of SDAM, MUTR \cite{mutr}, and ReferFormer \cite{referformer} on the Ref-YouTube-VOS dataset. Specifically, we randomly sample six identical text-video pairs from the Ref-YouTube-VOS dataset, plot the \(\mathcal{J\&F}\) metric of each frame in the video as a scatter plot, and use a Moving Average (MA) to draw the temporal stability trend lines for each method. The experimental result shows that our SD mechanism effectively improves temporal stability in sequence.

% figure: qualitative results
\begin{figure}[t]
\centering
\includegraphics[width=\linewidth]{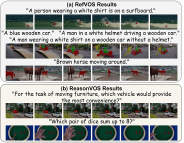}
\caption{Qualitative results on the RefVOS (a) and ReasonVOS (b) datasets. The time steps are directed from left to right. Zoom in for better viewing.}
\label{fig_qualitative}
\end{figure}

\subsection{Qualitative Results}
As shown in Figure \ref{fig_qualitative}, our method demonstrates excellent performance when facing challenging scenes in the RefVOS and ReasonVOS datasets. Our method performs excellently in challenging scenarios in RefVOS, including cases with small target objects, closely spaced similar objects, and complex prompts containing motion information (as shown in Figure \ref{fig_qualitative} (a)). Notably, the examples on the ReasonVOS dataset are impressive, where our method accurately infers the object described by the complex text query in scenarios with multiple similar objects and occlusions (as shown in Figure \ref{fig_qualitative} for the ReasonVOS). Our method possesses remarkable real-world reasoning abilities, enabling simple mathematical derivations based on information from the video frames (as shown in Figure \ref{fig_qualitative} for the ReVOS).

%================================ Conclusion================================%
\section{Conclusion}
In this paper, we have proposed novel SDAM, a Spatio-temporal Decoupling Training-Free Reasoning Video Segmentation with Adaptive Object Memory. We introduce an Adaptive Object Memory module to memorize key objects based on motion cues in the video sequence, and leverage a Spatio-temporal Decoupling mechanism to enhance the temporal stability in reasoning video segmentation. We achieve state-of-the-art performance on three RefVOS datasets and two ReasonVOS datasets. In the future, we aim to improve the model's robustness in keyframe selection and the stability in temporal propagation to adapt to diverse scenarios.

%================================ Acknowledgements ================================%
\section{Acknowledgements}
This work is supported in part by the National Natural Science Foundation of China (Grant No. 62176172, 61672364); partially by the National Key Research and Development Program of China (Grant No. 2018YFA0701701); partially by the Natural Science Foundation of Jiangsu Province (Grant for Young Scholars, Grant No. BK20250789); and partially by Undergraduate Training Program for Innovation and Entrepreneurship, Soochow University (No. 2025C104).

\bibliography{aaai2026}

\end{document}